\def\year{2021}\relax
\documentclass[letterpaper]{article} % DO NOT CHANGE THIS
\usepackage{aaai21}  % DO NOT CHANGE THIS
\usepackage{times}  % DO NOT CHANGE THIS
\usepackage{helvet} % DO NOT CHANGE THIS
\usepackage{courier}  % DO NOT CHANGE THIS
\usepackage[hyphens]{url}  % DO NOT CHANGE THIS
\usepackage{graphicx} % DO NOT CHANGE THIS
\usepackage{graphicx}  % DO NOT CHANGE THIS
\usepackage{natbib}  % DO NOT CHANGE THIS AND DO NOT ADD ANY OPTIONS TO IT
\usepackage{caption} % DO NOT CHANGE THIS AND DO NOT ADD ANY OPTIONS TO IT
\title{Multi Layer Neural Networks as Replacement for Pooling Operations}
\author {
    Wolfgang Fuhl,\textsuperscript{\rm 1}
    Enkelejda Kasneci, \textsuperscript{\rm 1}\\
}
\affiliations {
    \textsuperscript{\rm 1} University T\"ubingen, Sand 14, 72076 T\"ubingen, Germany \\
    wolfgang.fuhl@uni-tuebingen.de, enkelejda.kasneci@uni-tuebingen.de
}

\begin{document}

\maketitle

\begin{abstract}
Pooling operations, which can be calculated at low cost and serve as a linear or nonlinear transfer function for data reduction, are found in almost every modern neural network. Countless modern approaches have already tackled replacing the common maximum value selection and mean value operations, not to mention providing a function that allows different functions to be selected through changing parameters. Additional neural networks are used to estimate the parameters of these pooling functions.Consequently, pooling layers may require supplementary parameters to increase the complexity of the whole model. In this work, we show that one perceptron can already be used effectively as a pooling operation without increasing the complexity of the model. This kind of pooling allows for the integration of multi-layer neural networks directly into a model as a pooling operation by restructuring the data and, as a result, learnin complex pooling operations. We compare our approach to tensor convolution with strides as a pooling operation and show that our approach is both effective and reduces complexity. The restructuring of the data in combination with multiple perceptrons allows for our approach to be used for upscaling, which can then be utilized for transposed convolutions in semantic segmentation.
\end{abstract}

\section{Introduction}
Convolutional neural networks are the successor in many visual recognition tasks~\cite{krizhevsky2012imagenet,yuan2019object,ICCVW2019FuhlW,CAIP2019FuhlW,ICCVW2018FuhlW,NNETRA2020,UMUAI2020FUHL} as well as graph classification~\cite{zhao2019learning,orsini2015graph} and time series annotation~\cite{palaz2015analysis,connor1994recurrent}. The main focus of modern research on CNNs includes architecture improvements~\cite{he2016deep,howard2017mobilenets}, optimizer enhancements~\cite{kingma2014adam, qian1999momentum}, computational cost reduction~\cite{rastegari2016xnor,AAAIFuhlW}, validation~\cite{ICMV2019FuhlW}, training procedures~\cite{goodfellow2014generative}, and also building blocks like convolutions~\cite{long2015fully}, graph kernels~\cite{yanardag2015deep} or pooling operations~\cite{kobayashi2019global,kobayashi2019gaussian,eom2018alpha}. The aforementioned pooling operations are used for data reduction, reducing calculation costs and making the model robust against input variations. This is especially useful in applications like eye tracking~\cite{WF042019} where the computational resources are limited and there is a plethora of information which can be extracted from the eye movements~\cite{FCDGR2020FUHL,fuhl2018simarxiv,ICMIW2019FuhlW1,ICMIW2019FuhlW2,EPIC2018FuhlW,C2019,FFAO2019}. In addition, the used algorithms have to be as efficient as possible to ensure a high battery runtime~\cite{WTCDAHKSE122016,WTCDOWE052017,WDTTWE062018,VECETRA2020,CORR2017FuhlW1,ETRA2018FuhlW,WTDTWE092016,WTDTE022017,WTE032017}.

The pooling operation itself is inspired by the biological viewpoint of the visual cortex, based on a neuroscientific study~\cite{hubel1962receptive}. Most works suggest that max pooling is considered, biologically, to be the best operator~\cite{riesenhuber1998just,riesenhuber1999hierarchical,serre2010neuromorphic}. In practice, however, average pooling also works for CNNs, just as effectively as the combined approaches of max and average pooling. Based on this evidence, it can be surmised that the optimal pooling operation is dependent on the model, the task and the data set used. To further improve the accuracy of CNNs, simple pooling operations (e.g. max and average) are replaced by other static functions as well as by trainable operators.

The first group of operations is motivated by image scaling and uses wavelets~\cite{mallat1989theory} in wavelet pooling~\cite{williams2018wavelet} or other image scaling techniques~\cite{weber2016rapid} such as detailed-preserving pooling (DPP)~\cite{saeedan2018detail}. Another approach is the integration of formulas that can choose between several static pooling operations like max or average pooling. The first studies in this area focus on mixed pooling and gated pooling~\cite{lee2016generalizing,yu2014mixed}. These selective methods have been extended with parameterizable functions that can map many different average and max pooling operations, including learned norm~\cite{gulcehre2014learned} alpha~\cite{simon2017generalized}, and alpha integration pooling~\cite{eom2018alpha}. The approach was further refined according to the maximum entropy principle~\cite{kobayashi2019global,lee2016generalizing} and, as with alpha integration pooling~\cite{eom2018alpha}, equipped with parameters that can be trained and optimized in an end-to-end fashion. The global-feature guided pooling~\cite{kobayashi2019global} uses the input feature map to adapt pooling parameters. As a result, an additional CNN was used and jointly trained. In \cite{lee2016generalizing}, the authors proposed mixed max average pooling, gated max average pooling, and tree pooling. 

\begin{figure*}[h]
	\centering
	\includegraphics[width=0.99\textwidth]{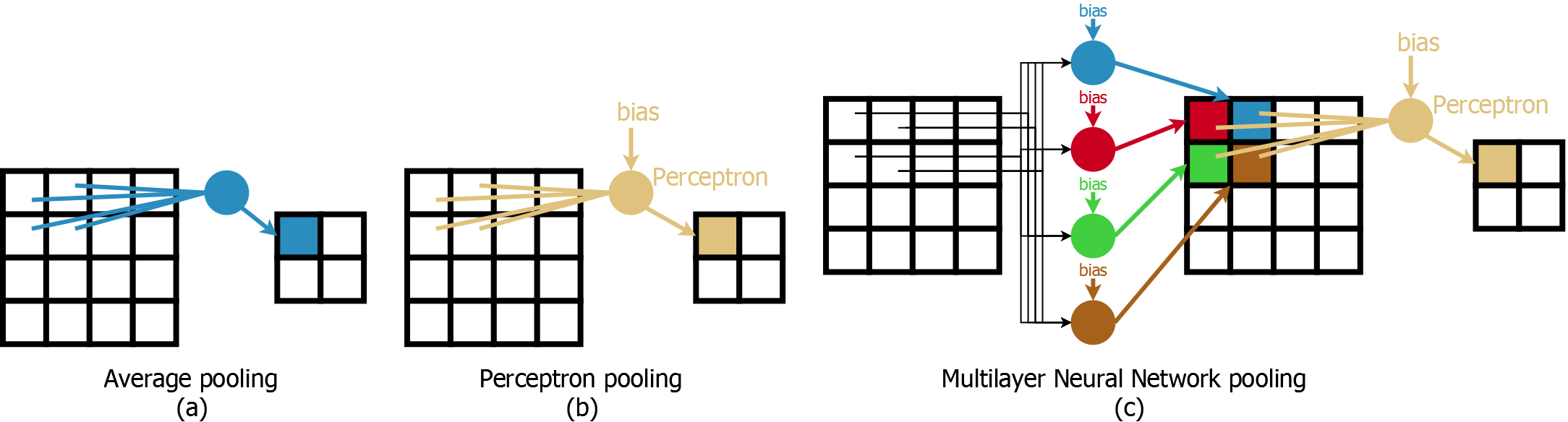}
	\caption{Visual explanation of our approach without activation functions. (a) is the average pooling operation where all weights are fixed to $0.25$. (b) is the simplest form of our approach which utilizes a perceptron as pooling operator. (c) is a multilayer neural network where the first (hidden) layer has four perceptrons and the output layer has one perceptron.}
	\label{fig:explain}
\end{figure*}

In addition to the deterministic pooling operations already mentioned, other methods that introduce randomness have been presented~\cite{zeiler2013stochastic}. The motivation of these pooling operations comes from drop out~\cite{srivastava2014dropout} and variational drop out~\cite{kingma2015variational}. This approach can also be used in combination with all other pooling operations. Another approach which does not formulate the combination of local neuron activations as a convex mapping or downscaling operation is gaussian based pooling~\cite{kobayashi2019gaussian}. The authors introduce a local gaussian probabilistic model with mean and standard deviation. The deviation are estimated using global feature guided pooling~\cite{kobayashi2019global} and, therefore, also require an additional CNN model for parameter estimation. Alternatives to those approaches include the commonly used strided tensor convolutions~\cite{springenberg2014striving}. Strided tensor convolutions require multiple parameters and network in network~\cite{lin2013network} wherein a small multilayer perceptron is used as convolution operation. Those multilayer perceptron convolutions are stacked like normal convolution layers but do not use any data resizing. In the end, the convolutions use one global average pooling as a downscaling operation before the fully connected layers.

In contrast to other approaches, we present the simple use of perceptrons~\cite{rosenblatt1958perceptron} or neurons as the pooling operator. To create a deeper network from these single neurons, we propose data restructuring, allowing the data to scale up. The pooling operation that we present can be be used not only in data reduction, but also in data expansion, a key element for semantic segmentation. By the simple use of neurons or multi-layer neural networks, there is only a minimal increase in the number of parameters in need of training and the complexity of the pooling operation remains nearly the same. In comparison to the other pooling operations presented, we also compare our approach to strided tensor convolutions~\cite{springenberg2014striving}. 

Our work contributes the latest research in the field with respect to the following points:
\begin{description}
	\item[1] We present an efficient usage of perceptrons as pooling operation and show a
	\item[2] Perceptron-based data upscaling.
	\item[3] We provide an efficient construction of multilayer neural networks with the proposed perceptron upscaling and perceptron pooling operations and
	\item[4] Provide CUDA implementations of the proposed approach for easy integration into research and application projects.
\end{description}

\section{Method}

Our fundamental idea to improve learnable pooling operations is to use one of the best known function approximators available today, i.e. the neural network which consists of single neurons (also called perceptrons) and is also known as multilayer perceptron (MLP). The main advantage of an MLP is that it can be easily integrated into deep neural networks (DNNs) since it consists of the same basic components as a DNN. This makes it easy to train with the remaining layers and the same optimization methods.

\begin{figure*}[h]
	\centering
	\includegraphics[width=0.99\textwidth]{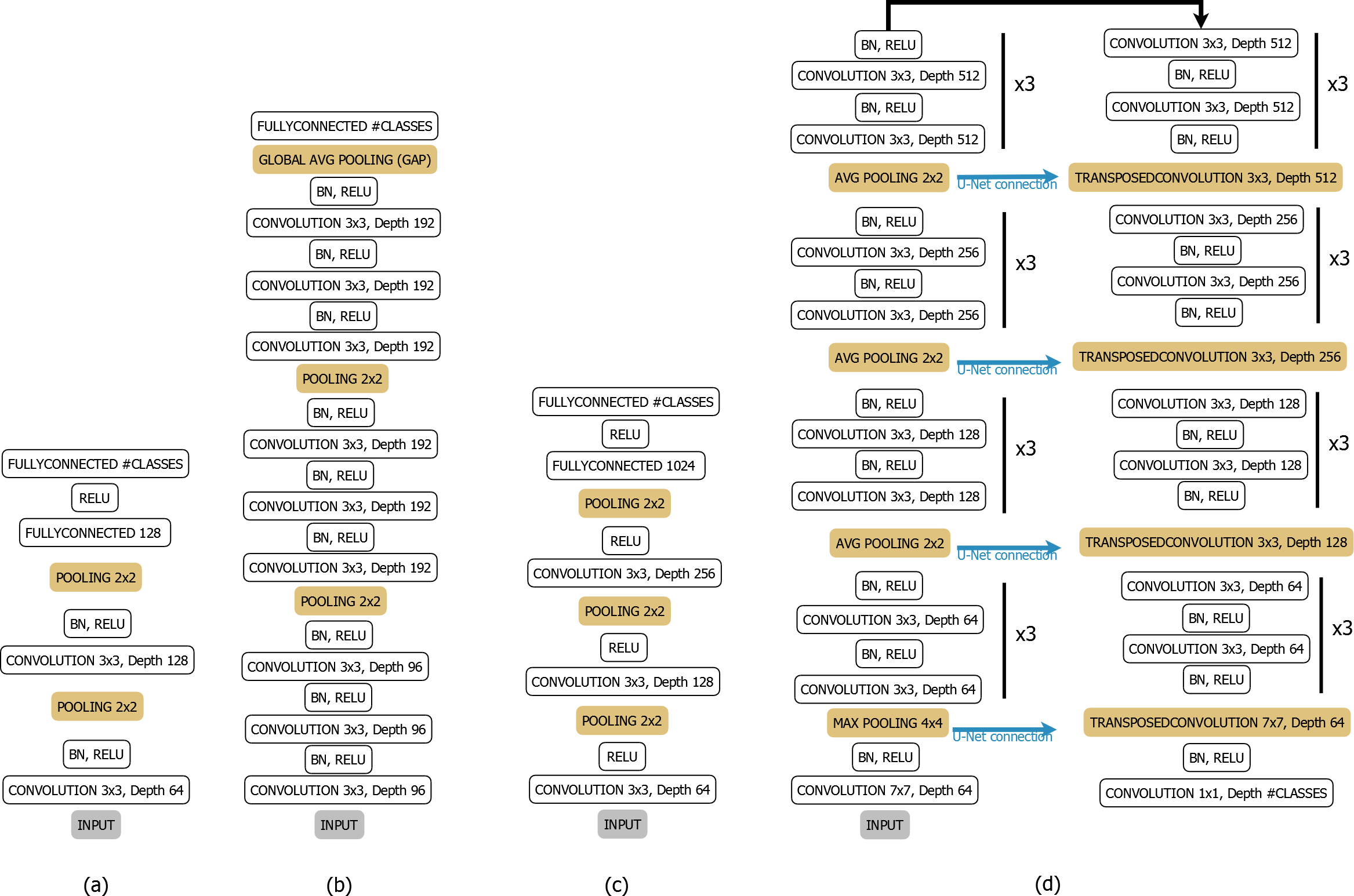}
	\caption{All used architectures in our experimental evaluation. The orange blocks are replaced with different pooling operations or in case of the transposed convolutions, the upscaling is replaced with our approach. (a) represents a small neural network model with batch normalization taken from \cite{eom2018alpha}. (b) is a 14 layer architecture taken from \cite{kobayashi2019gaussian}. (c) is a small model without batch normalization. (d) is a residual network using the interconnections from U-Net~\cite{ronneberger2015u} for semantic image segmentation.}
	\label{fig:models}
\end{figure*}

Figure~\ref{fig:explain} a) shows the basic concept of a pooling operation. Based on the input window, an output value is calculated, which differs depending on the selected pooling operation. Then the window is moved in the x and y dimension based on the stride parameter. If the pooling operation is average pooling, the weights (represented by the blue lines in Figure~\ref{fig:explain} a)) could be assigned the value $0.25$. Starting from here, it is easy to replace the pooling operation with a perceptron, since the only missing piece is the bias term (Figure~\ref{fig:explain} b)). The calculation of the output is nearly identical to the average pooling with the constant $0.25$ weights, which are multiplied by the corresponding input values. Afterwards, the sum is calculated together with the bias term and the activation function (ReLu~\cite{hahnloser2000digital,glorot2011deep}, Sigmoid, TanH, etc.) of the perceptron is computed. Now we have a perceptron which is used as a pooling operation. To create a multilayer neural network, we simply use several perceptrons with activation function in the first layer and attach further perceptrons to their outputs. This idea is shown in Figure~\ref{fig:explain} c), where four perceptrons are defined for the input window of $2 \times 2$ and their output is arranged in the x,y plane. For four perceptrons and a stride of two, the input tensor has the same size as the output tensor (see Figure~\ref{fig:explain} c)). One additional layer is then added on the arranged output of the four perceptrons. For the example shown in Figure~\ref{fig:explain} c), this layer consists of a perceptron with a stride of two and a window size of $2 \times 2$. Thus, we have defined a neural network with a hidden layer of 4 perceptrons and an output layer of 1 perceptron, which represents our pooling operation. 

The training of the perceptron or neural networks is the same as in any other layer of the superordinate neural network. As an additional memory requirement, the generated error is added, as in any other layer, to store the backpropagated error, which is needed to calculate the gradient. The only difference to the other layers in the neural network is that the learning rate of the perceptrons for the weights and the bias term should be reduced ($10^{-1}$ in our experiments). Weight decay can be used with the same reduction, but we disabled it to achieve slightly better results (Factor 0 in our experiments). It is, of course, also possible to train the perceptron or neural network for pooling at the same learning rate. In the case of large input and output tensors, however, the training becomes unstable. This is due to the fact that the error of the entire tensor affects only a few weights, and, therefore, the weights vary greatly. For example, for the nets in Figure~\ref{fig:models} a) and c) it is possible to use the same learning rate without problems. In case of Figure~\ref{fig:models} b) and d), however, this can lead to a initially fluctuating training phase. 

A further refinement for the effective use of perceptrons or neural networks as pooling operators is the initialization of the parameters. Normally, formula 16 from \cite{glorot2010understanding} is used for the random initialization of the parameters, which we have also used for all other layers. In the case of perceptron or neural networks, however, this can easily lead to failure. The networks only have a few parameters and, in the case of an unfavorable initialization, may not be able to shift through the gradient to a suitable minima. A simple example is the use of a perceptron for pooling with the average pooling parameter initialization. This means that each weight is set to $0.25$ and the bias term is set to $0$. Following the training of the entire model, the results of our evaluations were significantly better when compared to average pooling (see Table~\ref{tbl:exp0}).
\begin{table*}[h]
	\centering
	\caption{Results of a single percptron with the parameter initialization from avarage pooling (each weight is set to $0.25$ and the bias term to $0$). The model a) form Figure~\ref{fig:models} was used and evaluated on the CIFAR10 data set.}
	\label{tbl:exp0}
	\begin{tabular}{lccccc}
		\textbf{Pooling method} & \textbf{Run 1} & \textbf{Run 2} & \textbf{Run 3} & \textbf{Run 4} & \textbf{Run 5}\\ \hline
		Averge & 84.82 & 84.61 & 84.96 & 85.01 & 84.58 \\
		\textit{(ours) Perceptron (ReLu)} & 84.92 & 84.53 & 84.82 & 85.07 & 84.41 \\
		\textit{(ours) Perceptron} & \textbf{85.94} & \textbf{85.76} & \textbf{85.92} & \textbf{86.16} & \textbf{86.06}  
	\end{tabular}
\end{table*}

\begin{table*}[h]
	\centering
	\caption{Results of different pooling operations for model a) from Figure~\ref{fig:models} on the CIFAR10 data set. Our approaches are highlighted in italics. Perceptron means only a single perceptron for pooling. NN-4-1 is a multilayer neural network with 4 neurons in the first layer and 1 output neuron. NN-Z corresponds to one perceptron for pooling per layer of the input tensor. In NN-Field we used one perceptron per pooling region in the x,y plane and with NN-Tensor we used for each pooling region in the input tensor a seperate perceptron. ReLu is here the abbreviation for rectifier linear unit.}
	\label{tbl:exp1}
	\begin{tabular}{lcc}
		\textbf{Pooling method} & \textbf{Accuracy on CIFAR10} & \textbf{Additional Parameters}\\ \hline
		Average & 85.04 & 0 \\
		Max & 84.43 & 0 \\
		Strided tensor convolution (ReLu)~\cite{springenberg2014striving} & 87.70 & 82,112 \\
		Strided tensor convolution~\cite{springenberg2014striving} & 86.78 & 82,112 \\
		\textit{(ours) Perceptron (ReLu)} & 85.22 & 10 \\
		\textit{(ours) Perceptron} & \textbf{87.71} & 10 \\
		\textit{(ours) Perceptron no bias (ReLu)} & 84.71 & 8 \\
		\textit{(ours) Perceptron no bias} & 85.11 & 8 \\
		\textit{(ours) NN-4-ReLu-1-ReLu}& 85.45 & 50 \\
		\textit{(ours) NN-4-ReLu-1}& 86.40 & 50 \\
		\textit{(ours) NN-4-1}& 87.29 & 50 \\
		\textit{(ours) NN-Z (ReLu)}& 83.87 & 770 \\
		\textit{(ours) NN-Z}& 84.37 & 770 \\
		\textit{(ours) NN-Field (ReLu)}& 84.23 & 1,600\\
		\textit{(ours) NN-Field}& 85.28 & 1,600\\
		\textit{(ours) NN-Tensor (ReLu)}& 81.04 & 122,880 \\
		\textit{(ours) NN-Tensor}& 80.93 & 122,880
	\end{tabular}
\end{table*}

Table~\ref{tbl:exp0} shows that the ReLu has a considerable influence on the result. We will explore this in greater detail in the first Experiment~\ref{sec:exp1}. Consequently, for our initialization we have also used random values, but ensured that they are either symmetrical or follow rotated/mirrored patterns based on the sign of the values. This concept stems from manually created filters that originate in classical image processing, such as edge filters and weighted average pooling. In the case of a single perceptron this means: either all positive or negative and vice versa, a diagonal negative and the rest positive, or that the transition between positive and negative  is along the x or y axis. For several perceptrons in the same layer, we calculated a random pattern and rotated it or mirrored it along the x or y axis, making sure that the pattern was not repeated. This was repeated until all perceptrons in the layer had an initialization.

%params
\textbf{Additional parameters for a perceptron: } Each perceptron or neuron has a input window size of $W \times H$ and a bias term $b$. Therefore, we have $W*H+1$ additional parameters for a perceptron. 

\textbf{Additional parameters for the multilayer neural network:} The amount of neurons in the first layer is $p_{1}$ and in the last layer $p_{L}$. For the first layer we would have $p_{1}*(W_{1}*H_{1}+1)$ additional parameters. Each following layer has $p_{l}*(W_{l}*H_{l}+1)$ parameters. Thus, the total number of parameters can be specified as $\sum_{l=1}^{L}p_{l}*(W_{l}*H_{l}+1)$.

\textbf{Complexity of the perceptron:} We perform per index value one multiplication and one addition. For the bias term we need an extra addition. Therefore, the complexity is $O(\frac{2*W*H*n}{stride^2}+\frac{n}{stride^2})$ which is theoretical $O(n)$ as for the standard pooling operations.

\textbf{Complexity of the multi layer neural network: } The amount of neurons in the first layer is $p_{1}$ and in the last layer $p_{L}$. Furthermore, we have $n_{l}$ input values at layer $l$. Therefore, the first layer needs $O(\frac{p_{1}*2*W_{1}*H_{1}*n_{1}}{stride_{1}^2}+\frac{n_{1}}{stride_{1}^2}))$ operations. The following layers need $O(\frac{p_{l}*2*W_{l}*H_{l}*n_{l}}{stride_{l}^2}+\frac{n_{l}}{stride_{l}^2}))$ operations. Since the amount of perceptrons or neurons per layer is independent of $n$ we still have a theoretical complexity of $O(n)$. With $stride_{l}^2$ we expect the same shift in each x and y dimension of the input tensor at layer $l$.

\section{Neural Network Models}

\begin{table*}[h]
	\centering
	\caption{Results of different pooling operations for model b) form Figure~\ref{fig:models} on the CIFAR100 data set. Our approaches are highlighted in italics. Perceptron means only a single perceptron for pooling. NN-4-1 is a multilayer neural network with 4 neurons in the first layer and 1 output neuron. The same notation was used for NN-16-1 with 16 neurons in the first layer. In the last entry we also replaced the GAP layer with a perceptron.}
	\label{tbl:exp2}
	\begin{tabular}{lcc}
		\textbf{Pooling method} & \textbf{Accuracy on CIFAR100} & \textbf{Additional Parameters}\\ \hline
		Average &  75.40 & 0 \\
		Max &  75.36 & 0 \\
		Strided tensor convolution (ReLu)~\cite{springenberg2014striving} & \textbf{77.53} & 184,608 \\
		Stochastic~\cite{zeiler2013stochastic} &  75.66 & 0 \\
		Mixed~\cite{lee2016generalizing} &  75.90 & 2 \\
		DPP~\cite{saeedan2018detail} &  75.56 & 4 \\
		Gated~\cite{lee2016generalizing} &  76.03 & 18 \\
		GFGP~\cite{kobayashi2019global} &  75.81 & 46,080 \\
		Half-Gauss~\cite{kobayashi2019gaussian} &  76.74 & 69,840 \\
		iSP-Gauss~\cite{kobayashi2019gaussian} &  76.85 & 69,840\\
		\textit{(ours) Perceptron}& 76.06  & 10 \\
		\textit{(ours) NN-4-1}& 76.21 & 50 \\
		\textit{(ours) NN-16-1}& 77.14 & 194 \\
		\textit{(ours) Perceptron \& GAP}& 76.37  & 75 
	\end{tabular}
\end{table*}

Figure~\ref{fig:models} shows all the architectures used in our experiments. Figure~\ref{fig:models} a) shows a small neural network we adapted from \cite{eom2018alpha} and is employed in Experiment 1 to compare different pooling operations as well as spatial pooling with fields and tensors of neurons on the CIFAR10 data set~\cite{krizhevsky2009learning}. The network in Figure~\ref{fig:models} b) was taken over from \cite{kobayashi2019gaussian} and is used for comparison with the state-oft-the-art on the CIFAR100 data set~\cite{krizhevsky2009learning} as shown in Experiment 2. The third model (Figure~\ref{fig:models} c)) is used in Experiment 3 and does not include batch normalization. This model was employed to compare pooling operations with the same random initialization and the same batches during training. The last model in Figure~\ref{fig:models} d) is a fully convolutional neural network~\cite{long2015fully} with U-Net connections~\cite{ronneberger2015u}. It is used to compare the pooling operations and the high scaling for semantic segmentation. We implemented our approach into DLIB~\cite{king2009dlib} and also used it for all evaluations and comparisons.

\section{Datasets}
In this section, we present and explain training parameters for all the datasets used in our experiments. We also define the batch size as well as the optimizer and its parameters. In the case of data augmentation, we kept the number of datasets to a minimum for reproduction purposes, described in detail in the following section.

\textbf{CIFAR10}~\cite{krizhevsky2009learning} consists of 60,000 $32 \times 32$ colour images. The dataset has ten classes. For training, 50,000 images are provided with 5.000 examples in each class. For validation, 10,000 images are provided (1,000 examples for each class). The task in this dataset is to classify a given image to one of the ten categories.

\textit{\textbf{Training:} We used a batch size of 50 with a balanced amount of classes per batch and an initial learning rate of $10^{-3}$. As optimizer, we used ADAM~\cite{kingma2014adam} with weight deacay of $5*10^{-5}$, momentum one with $0.9$ and momentum two with $0.999$. For data augmentation, we cropped a $32 \times 32$ region from a $40 \times 40$ image, where the original image was centered on the $40 \times 40$ image and the border on each side are 4 pixels set to zero. The training itself was conducted for 300 epochs, whereby the learning rate was decreased by $10^{-1}$ after each 50 epochs. The images are preprocessed by mean substraction (mean-red $122.782$, mean-green $117.001$, mean-blue $104.298$) and division by $256.0$.}

\textbf{CIFAR100}~\cite{krizhevsky2009learning} is similar to CIFAR10 and consists of 32x32 color images, which must be assigned to one out of 100 classes. For training, 500 examples of each class are provided. The validation set consists of 100 examples for each class. Thus, CIFAR100 has the same size as CIFAR10, with 50,000 images in the training set and 10,000 images in the validation set, respectively.

\textit{\textbf{Training:} We used a batch size of 100 and an initial learning rate of $10^{-1}$. As optimizer we used SGD with momentum~\cite{qian1999momentum} ($0.9$) and a weight decay of ($5*10^{-4}$). For data augmentation, we normalized the images to zero mean and one standard deviation and cropped a $32 \times 32$ region from a $40 \times 40$ image, where the original image was centered on the $40 \times 40$ image and the border on each side are 4 pixels set to zero. The training itself was conducted for 160 epochs, whereby after the 80th and 120th epoch the learning rate was decreased by $10^{-1}$. This is the same procedure as specified in \cite{kobayashi2019gaussian}.}

\textbf{VOC2012}~\cite{pascalvoc2012} is a detection, classification and semantic segmentation dataset. We only used the semantic segmentations in our evaluation, which contains 20 classes. The task for semantic segmentation is to provide a pixelwise classification of a given image. Each image can contain multiple objects of the same class, but not all classes are present in each image.  Therefore, the amount of classes increases to 21. For training, 1,464 images are provided with a total of 3,507 segmented objects. For validation, another 1,449 images are designated with a total of 3,422 segmented objects. In this dataset, the number of objects is unbalanced, making the dataset more challenging to utilize. In addition to the training and validation set's segmented images a third set without segmentations is provided, containing 2,913 images with 6,929 objects. We did not use the third dataset in our training.

\begin{table*}[h]
	\centering
	\caption{Results of different pooling operations for model c) form Figure~\ref{fig:models} on the CIFAR10 data set. Our approaches are highlighted in italics. Perceptron mean only a single perceptron for pooling. NN-4-1 is a multilayer neural network with 4 neurons in the first layer and 1 output neuron. Each convolution and fully connected layer had the same random initialization as well and all models saw the same batches during training.}
	\label{tbl:exp3}
	\begin{tabular}{lccccc}
		\textbf{Pooling method} & \textbf{Run 1} & \textbf{Run 2} & \textbf{Run 3} & \textbf{Run 4} & \textbf{Additional Parameters}\\ \hline
		Averge & 84.12 & 84.13 & 84.23 & 84.47 & 0 \\
		Max & 85.63 & 85.77 & 85.36 & 86.01 & 0 \\
		Strided tensor convolution (ReLu)~\cite{springenberg2014striving} & \textbf{86.95} & \textbf{87.68} & 87.11 & 87.84 & 344,512 \\
		\textit{(ours) Perceptron} & 85.73 & 86.13 & 85.95 & 85.18 & 15 \\
		\textit{(ours) NN-4-1} & 86.37 & 87.15 & \textbf{87.21} & \textbf{87.89} & 75 
	\end{tabular}
\end{table*}

\begin{table*}[h]
	\centering
	\caption{Average pixel classification accuracy on Pascal VOC2012 sematic segmentation dataset with model d) from Figure~\ref{fig:models}. Our approaches are highlighted in italics. Perceptron is the downscaling operation (One single perceptron) and NN-4/16-UP are four/sixteen neurons for upscaling. The sixteen neurons are in the last layer before the output.}
	\label{tbl:exp4}
	\begin{tabular}{lcc}
		\textbf{Pooling method} & \textbf{Pixel accuracy on VOC2012} & \textbf{Additional Parameters} \\ \hline
		As in Figure~\ref{fig:models} d) & 85.15 & 0 \\
		\textit{(ours) Perceptron} \& Transpose  & 86.36 & 32 \\
		\textit{(ours) Perceptron \& NN-4/16-UP} & \textbf{87.62} & 172
	\end{tabular}
\end{table*}

\textit{\textbf{Training:} We used a batch size of 10 and an initial learning rate of $10^{-1}$. As optimizer we used SGD with momentum~\cite{qian1999momentum} ($0.9$) and weight deacay ($1*10^{-4}$). For data augmentation, we used random cropping of $227 \times 227$ regions with a random color offset and left right flipping of the image. The training itself was conducted for 800 epochs, whereby after each 200 epochs the learning rate was decreased by $10^{-1}$. The images are preprocessed by mean substraction (mean-red $122.782$, mean-green $117.001$, mean-blue $104.298$) and division by $256.0$.}

\section{Experiment 1: Spatial Invariant vs Spatial Pooling}
\label{sec:exp1}
Table~\ref{tbl:exp1} shows the comparison of different pooling operations on the CIFAR10 data set. The model chosen was a) from Figure~\ref{fig:models}. Each pooling operation was trained a total of ten times with random initialization. Of all ten runs, the best result was entered in Table~\ref{tbl:exp1}. First, Table~\ref{tbl:exp1} shows that a single perceptron as a pooling operation is as good as a tensor convolution with stride. Additionally, from the Table~\ref{tbl:exp0} one can see that a multi-layer neural network (NN-4-1) does not perform as well. The single perceptron was also trained and evaluated without bias term and, as exhibited, it performed only slightly better than average pooling. Thus, it can be assumed that the bias term has a significant influence on this model and data set.

Another clear observation that can be obtained from this evaluation is that the ReLu (Rectifier Linear Unit) has a strong limiting influence on the classification accuracy. We believe this is the case because we used the neural network like a function embedded in a larger network. By restricting the network, we reduce the amount of functions that can be learned. Similar to a directly used neural network, the outputs are not limited. As the tiny neural networks with ReLu score significantly worse in all evaluations, we do not use the ReLu in the following experiment. For the strided tensor convolution~\cite{springenberg2014striving} as pooling operation we continued with the ReLu due to better results.

As in \cite{lee2016generalizing} we have additionally evaluated spatially separated placements of neurons (NN-Z, NN-Field, and NN-Tensor). NN-Z is a separate perceptron for each channel of the input tensor. For the NN-Field, we assigned a single perceptron to all pooling windows in the x,y plane and moved them along the channels. In the last evaluated spatial arrangement NN-Tensor, we assigned a single perceptron to each pooling region in the input sensor. As can be seen in Table~\ref{tbl:exp1}, the accuracy of all is significantly worse than the standard max and average pooling operations. The worst is NN-Tensor, which requires more parameters than the strided tensor convolution~\cite{springenberg2014striving}. Thus, we can confirm for the perceptrons that a spatial arrangement does not provide any improvement, as the authors in \cite{lee2016generalizing} have confirmed in their approach.

\section{Experiment 2: Comparison to the state-of-the-art}
Table~\ref{tbl:exp2} shows the comparison of our approach with the state-of-the-art on the CIFAR100 data set. As in \cite{kobayashi2019gaussian}, we have trained each model three times with random initialization. In the end, we entered the best results in Table 2. As can be seen, the strided tensor convolution~\cite{springenberg2014striving} has achieved the best results, but it also requires the most additional parameters (184,608). The second best results are obtained with the NN-16-1 neural network (97additional parameters), the iSP-Gauss~\cite{kobayashi2019gaussian} (69,840 additional parameters) and the Half-Gauss~\cite{kobayashi2019gaussian} (69,840 additional parameters). This is followed by the our two smaller models with a single perceptron and tiny neural network which both require significantly less additional parameters, i.e., only 50,  compared to the above mentioned Gaussian-based approaches. If the global average pooling (GAP) is replaced by a perceptron, the number of parameters increases by 65 and the result improves by 0.34\%. To perform training with the perceptron as a GAP replacement, we have set the learning rate factor (bias and weights) for this perceptron to $10^{-3}$. At this point, it must also be mentioned that our approach can be calculated in O(n) and we have only evaluated very small neural networks. It is, of course, also possible to use deeper and wider nets as pooling operations.

\section{Experiment 3: Equal Randomness and Batch Data Comparison}

Table~\ref{tbl:exp3} shows an evaluation of different pooling operations, where the same initial parameters of the convolution layers and fully connected layers are set for all. The data set used is CIFAR10 and the model is c) from Figure~\ref{fig:models}. Of course, this does not apply to the parameters of the pooling operations because of their different sizes. Also, the individual batches and the sequence of the batches were the same for all models. In this evaluation, we wanted to show a comparison between pooling operations under the same conditions. As can be seen in Table~\ref{tbl:exp3}, the overall best result was achieved by the NN-4-1 in the fourth evaluation. Comparing the NN-4-1 with the tensor convolution, the results are always similar, whereas the tensor convolution is more stable in its range of values. A closer look at the standard pooling operations' max and average pooling reveals that max pooling is consistently better than average pooling for this data set with the model c) from Figure~\ref{fig:models}. If we compare the individual perceptron with max and average pooling, it outperforms both in three of four runs for the model c) from Figure~\ref{fig:models} and the CIFAR10 data set.

\section{Experiment 4: Usage in Semantic Segmentation}

Table~\ref{tbl:exp4} shows the result of the U-Net from Figure~\ref{fig:models} d) on the VOC2012 data set. Each net was initialized and trained with random values. For Perceptron \& Transopse we replaced only the pooling operations with a perceptron. For Perceptron \& NN-4/16-UP we replaced the pooling and upscaling operations with perceptrons. As can be seen, our approach improves the results both as a pooling operation and for up scaling. Since VOC2012 is a very hard data set and semantic segmentation is a difficult task, we see this as a significant improvement of the results.

\section{Limitations}
Despite the parameter reduction presented above, our methods still have some disadvantages when compared to the classical maximum value selection or the mean value pooling. One disadvantage is that we still have a few additional parameters to calculate for the perceptron or the neural network. Additionally, this means that we have to provide memory for back-propagating the error, as is the case for each learning layer in a neural network. Of course, this also affects the optimizer, which requires needs additional memory for the moments. The use of neural networks as pooling operators also extends to the search space for model finding and, thus, their complexity and computing requirements. However, in general, our approach does not increase the complexity of the calculation of a pooling operation in the case of the perceptron, but it does improve the accuracy of the model. In the case of using a multilayer neural network for the pooling operation, our approach naturally increases the number of computations. When compared to a tensor convolution as pooling operation, however, the increase inherent in our approach is only minimal because the tensor convolution increases the complexity by the output tensor depth. As a general remark, it must also be said that, in case of unstable training, reducing the learning rate of the perceptron or small neural network has always resulted in success.

\section{Conclusion}
In this paper, we have shown that single perceptrons can be used effectively as pooling operators without increasing the complexity of the model. We have also shown that neural networks can be formed as pooling operators by simply restructuring the output data of several perceptrons. This increases the complexity and number of parameters in the model only minimally compared to tensor convolutions as pooling operator and is almost as effective. These multi-layer neural networks and presented restructuring can also be used to learn a scaling that can be effectively employed for transposed convolutions. Here it is also possible to learn the scaling via tensors. This would require an extension of our approach utilizing two dimensional matrices. In addition to the models evaluated in this paper, it is, of course, possible to train deeper nets as pooling operators or to equip individual layers with more perceptrons. In this way, the results can be further improved. We leave this open for future research. Our approach is easy to integrate into modern architectures and can be learned simultaneously with all other parameters without creating parallel branches in a model. Thus, the approach can also be effectively computed on a GPU.

\bibliographystyle{aaai21}
\bibliography{mybib}

\end{document}